\documentclass[10pt,twocolumn,letterpaper]{article}

\usepackage{iccv}
\usepackage{times}
\usepackage{epsfig}
\usepackage{graphicx}
\usepackage{amsmath}
\usepackage{amssymb}

\usepackage[breaklinks=true,bookmarks=false]{hyperref}

\newcommand{\G}{\mathcal{G}}

\newcommand{\U}{\mathcal{U}}

\newcommand{\F}{\mathcal F}

\newcommand{\A}{\mathcal A}

\newcommand{\N}{\mathcal N}

\newtheorem{proof*}{Proof}[section]

\newtheorem{definition*}{Definition}[section]

\iccvfinalcopy % *** Uncomment this line for the final submission

\def\iccvPaperID{****} % *** Enter the ICCV Paper ID here

\ificcvfinal\fi

\begin{document}

%%%%%%%%% TITLE
\title{AdaFNIO: Adaptive Fourier Neural Interpolation Operator for video frame interpolation}

\author{Hrishikesh Viswanath\\
Purdue University\\
{\tt\small hviswan@purdue.edu}
\and
Md Ashiqur Rahman\\
Purdue University\\
{\tt\small rahman79@purdue.edu}
\and
Rashmi Bhaskara\\
Purdue University\\
{\tt\small bhaskarr@purdue.edu}
\and
Aniket Bera\\
Purdue University\\
{\tt\small ab@cs.purdue.edu}
}

\maketitle
\ificcvfinal\thispagestyle{empty}\fi

%%%%%%%%% ABSTRACT
\begin{abstract}
   We present, \textbf{AdaFNIO} - Adaptive Fourier Neural Interpolation Operator, a neural operator-based architecture to perform video frame interpolation. Current deep learning-based methods rely on local convolutions for feature learning and suffer from not being scale-invariant, thus requiring training data to be augmented through random flipping and re-scaling. On the other hand, \textbf{AdaFNIO}, learns the features in the frames, independent of input resolution, through token mixing and global convolution in the Fourier space or the spectral domain by using Fast Fourier Transform (FFT). We show that \textbf{AdaFNIO} can produce visually smooth and accurate results. To evaluate the visual quality of our interpolated frames, we calculate the structural similarity index (SSIM) and Peak Signal to Noise Ratio (PSNR) between the generated frame and the ground truth frame. We provide the quantitative performance of our model on Vimeo-90K dataset, DAVIS, UCF101 and DISFA+ dataset. 
\end{abstract}

%%%%%%%%% BODY TEXT
\section{Introduction}

\begin{figure}[h!]
    \centering
    \includegraphics[width=8cm]{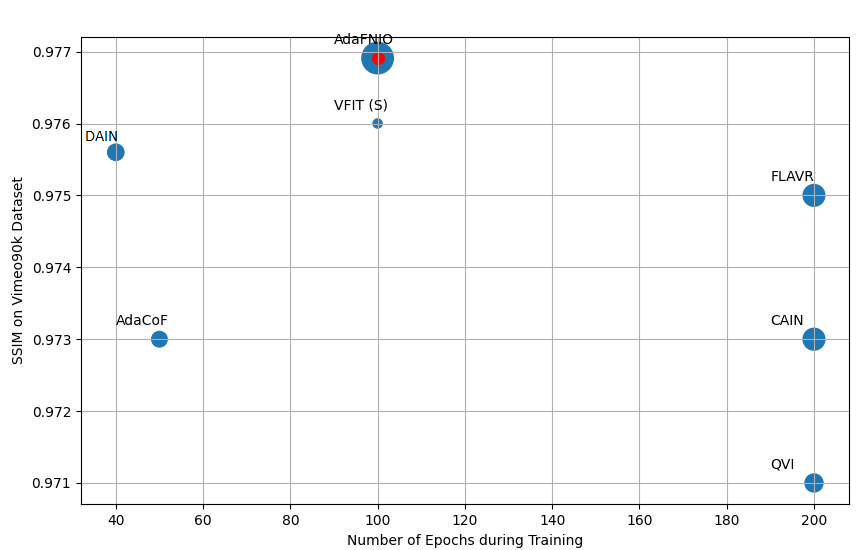}
    \caption{Comparison of Performance of Deep Learning based models on Vimeo90K dataset against the number of epochs. The proposed NIO model achieves comparable accuracy within just \textbf{10 epochs}.}
    \label{fig:my_label}
\end{figure}

Video frame interpolation is the process of creating several in-between frames from the set of available frames. This is a challenging problem as it involves understanding the geometric structures of images and predicting the positions of multiple objects within an image while taking into account the varying velocities of the objects and the time step of the frames. In addition, a video interpolation system should be easy to run on commonly used devices and be able to operate on edge hardware and work on videos of any arbitrary resolution. Once trained, neural networks offer a low-cost solution to interpolation since a device only needs to store the weights, which are a few hundred megabytes in size. However, neural networks that only rely on convolutional filters don't generalize well to scaling since the filters are of fixed size and recognize patterns that conform to the filter size.

Many applications make use of video frame interpolation, such as apps that generate movies and panoramic views from visually similar frames and applications that run on the network edge that may need to recover lost frames due to network issues, restore broken videos \cite{cheng2021audio, liang2022vrt,kim2018spatio}. Recent works focus on increasing the video frame rate to improve gaming and real-time video streaming experience. The other applications include medical imaging \cite{ali2021deep, karargyris2010three}, restoring compressed videos \cite{he2020video} and generating virtual frames in 3D spaces \cite{smolic2008intermediate, wang2010depth}. Most of these applications, especially video streaming applications and gaming environments, need the interpolation algorithm running on the network edge while also handling very high-resolution videos. Several cutting-edge models have been developed, and they produce interpolated frames that, on average, have a structural similarity of 98$\%$ with the expected output but are trained on small patches of input. While it is very expensive computationally to train neural networks on high-resolution inputs, it is possible to design architectures that can be trained on low-resolution inputs but generalize well to high-resolution ones. In this paper, we present a powerful neural operator-based architecture for interpolating video frames through token mixing that has a quasi-linear complexity and is independent of input resolution \cite{guibas2021adaptive}. 

Current deep learning-based interpolation models are generally composed of Convolutional layers \cite{niklaus2017video, niklaus2018context, cheng2020video}. These models construct the missing frames by extracting the structural information present in the input images using the appropriate filters or kernels. Convolutional layers exhibit shift invariance as they capture objects present in different regions of the image. However, they are not invariant to scale or rotation. To overcome this issue, images are randomly flipped and rotated to capture different orientations of the same object. Moreover, Convolution neural network-based models rely on local convolution for feature learning and require large amounts of training data and take a long time to converge. There have been attempts to solve this issue, for example, video interpolation using cyclic frame generation technique \cite{liu2019deep}, but it is not very accurate.

Optical flow-based techniques \cite{krishnamurthy1999frame}, which capture the motion of objects, have also been applied for frame interpolation.  In this technique, the apparent motion of the pixels is captured and labeled as a flow vector. Using the estimated values of these flow vectors for each pixel, a missing frame can be generated. Flow-based techniques resolve the limitations imposed by inadequate kernel sizes in CNN-based methods and the frames can also be generated at a higher frequency per second, resulting in a smoother video. Optical flow-based methods fail while dealing with noisy frames due to the lack of necessary pixel information. It has also been observed that a combination of both kernel-based and optical flow-based methods \cite{choi2020channel} with Deep learning techniques like transformers \cite{shi2022video} and GANs have low frame interpolation errors and also provide a great frame rate for a smooth video. However, GAN based architectures suffer from modal collapse and thus cannot be generalized in an arbitrary fashion. They rely heavily on the distribution of input and would require re-training for a new distribution if the input space is changed. Transformer-based architectures have been shown to be very efficient \cite{shi2022video}. However, due to their massively complex architecture, they require a lot of computing power and powerful GPUs to train. Long training times are an additional downside.

In this paper, we define the problem of video interpolation from a physics perspective. The problem can be defined as predicting the trajectories of objects, each moving with a different velocity in continuous space, similar to optical flow, but we present a way to capture the flow information efficiently using kernels. This problem is similar to predicting the trajectory of wind currents or ocean currents, for which neural operators \cite{kovachki2021neural, guibas2021adaptive, rahman2022u} have been shown to be exceptionally efficient, beating state-of-the-art weather prediction models \cite{pathak2022fourcastnet}.  We present a powerful Fourier Neural Operator-based architecture - AdaFNIO - Adaptive Fourier Neural Interpolation Operator and show empirically that the model achieves state-of-the-art results. We quantify the quality of the results using the Peak Signal to Noise Ratio (PSNR) and the Structural Similarity Index and show that the SSIM of the generated frames is structurally similar to the ground truth.

The AdaFNIO network is a novel extension to the AdaCoF network, which captures complex motions and overcomes degrees of freedom limitations imposed by typical convolutional neural networks. AdaFNIO imposes resolution invariance to the AdaCoF network through spectral convolution layers. The network has a sequence of spectral convolution layers, which perform convolution upon translating the input to the Fourier space or spectral domain. The low-frequency regions are retained and the high-frequency ones are discarded because high-frequency points are specific to the particular input and may lead to overfitting, as shown in \cite{li2020fourier}. The spectral convolution architecture of the AdaFNIO network is similar to an autoencoder network where the encoder layers contract the input space to capture the key information about the distribution of the input. The decoder expands it back to its original input space.

In this work, we have three main contributions:
\begin{itemize}
\item We present a powerful, efficient neural operator-based architecture to perform video frame interpolation whose performance is comparable to state-of-the-art interpolation models. To the best of our knowledge, AdaFNIO is the first to propose a resolution invariant architecture to solve this problem.
\item We leverage the fact that learning in the Fourier domain allows for resolution-independent learning and allows for generalization to high-resolution images to capture finer details in high-resolution images that are harder to capture. 
\item Lastly, we show that AdaFNIO can generalize well on the unseen data by testing it on DAVIS-90 \cite{Caelles_arXiv_2019}, UCS101 \cite{liu2017voxelflow} and DISFA+ datasets. \cite{mavadati2016extended}. \end{itemize}  

\section{Literature Review}
Existing video interpolation methods focus on creating techniques that can accurately capture an object's motion while accounting for occlusions and blurry frames. To deal with blurry frames, the pyramid module proposed in \cite{shen2020blurry} successfully restores the inputs and synthesizes the interpolated frame.  Many other methods focus on estimating the bidirectional optical flow in the input frames \cite{liu2020video, krishnamurthy1999frame, hu2022many, raket2012motion, niklaus2020softmax} and these methods are usually trained using deep learning models like neural networks \cite{huang2022real}, generative adversarial networks \cite{tran2022video, bharadwaj2021video}, long short term memory (LSTM) \cite{hu2021capturing} and autoencoders \cite{chen2017long, park2020video}. However, these methods fail to generate smooth interpolation of the frames when dealing with large motions and occlusion in the input frames.

Bao et al. \cite{bao2019depth} introduced depth aware flow projection layer that uses depth maps in combination with optical flow to suppress the effect of occluded objects. In addition to bidirectional optical flow estimation, \cite{niklaus2018context} uses pixel-wise contextual information to generate high-quality intermediate frames and also effectively handles occlusion. To capture non-linear motion among frames, a bidirectional pseudo-three-dimensional warping layer is used in \cite{luo2022bi} that uses motion estimation and depth-related occlusion estimation to learn the interpolation. ST-MFNet \cite{danier2022st} uses 3D CNNs, and spatio-temporal GANs to estimate intermediate flows to capture large and complex motions.

The optical flow estimating models are accurate, but their calculations are expensive, and their designs are complicated. An alternate technique was kernel-based, \cite{shi2021video, peleg2019net, tian2022video, liu2019deep}, which uses filters to learn features from the input frames in order to synthesize an intermediate frame. These models are end-to-end trainable but fail to capture motion and pixel information beyond the kernel size. To overcome these limitations and also to handle other major issues like occlusion Choi et al. \cite{choi2020channel} proposed an architecture that uses a layer called PixelShuffle. The PixelShuffle layer in \cite{choi2020channel} downsizes the input frames to capture relevant feature information and upscales the feature maps in later stages to generate the missing frame and is a replacement for flow-estimation networks. A similar model that uses transformers has been proposed in \cite{kim2022cross}, which uses a visual transformer module that analyzes input frame features to refine the interpolated prediction and also introduces an image attention module to deal with occlusion. To avoid the additional computation overheads of having an image attention module, \cite{shi2022video} uses transformers along with local attention, which is extended to a spatial-temporal domain to compute the interpolation. 
\begin{figure*}[h]
    \centering
    \includegraphics[width=0.9\textwidth]{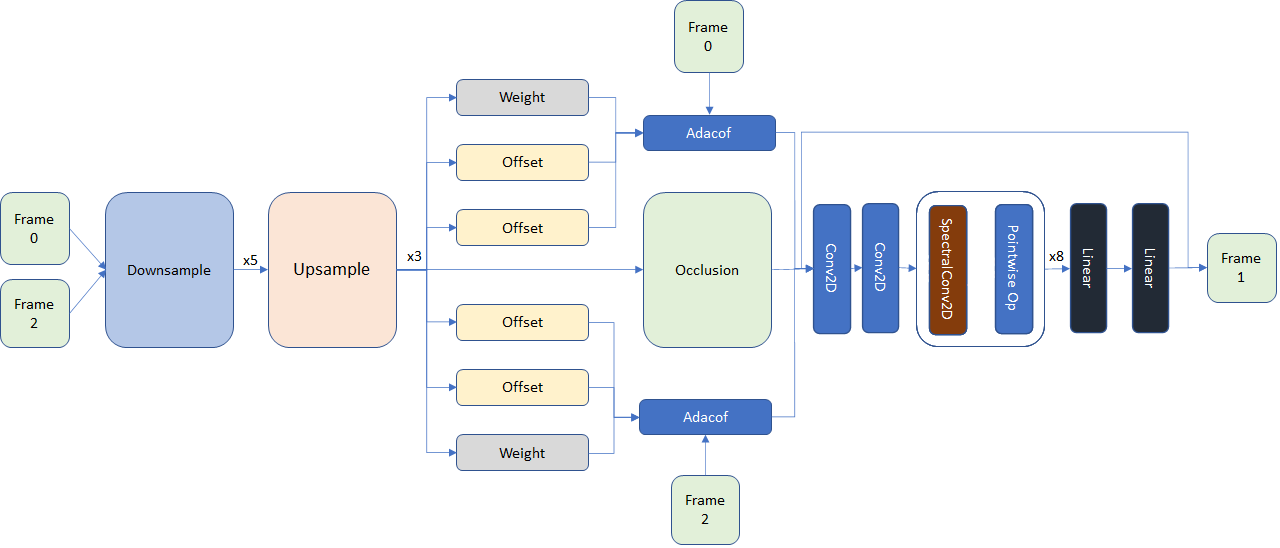}
    \caption{Abstract view of the ADAFNIO Architecture. Inspired by the AdaCoF Network, this architecture combines the AdaCoF network with neural operator layers. Initially, the two frames are fed to the encoder (Downsample) network, which applies successive 2D Convolution to the frames to generate a latent representation. The Upsample block consists of a sequence of upsample and convolution layers to reconstruct the frame from the latent representation. This reconstructed matrix is fed to two sets of 3 subnetworks - 2 Offset networks and 1 Weight network. These sub-networks extract the features for the AdaCof layer, which processes these features with the input frames. The final output is fed to the neural operator layers to extract finer information not captured by AdaCoF. The SpectralConv layers perform pixel-wise multiplication in the Fourier domain by first performing FFT on the frames. This process is analogous to token mixing. The final output is a weighted sum of the AdaCoF output and the Neural Operator output}
    \label{fig:overall_arch}
\end{figure*}
\begin{figure}[h]
    \centering
    \includegraphics[width=8cm]{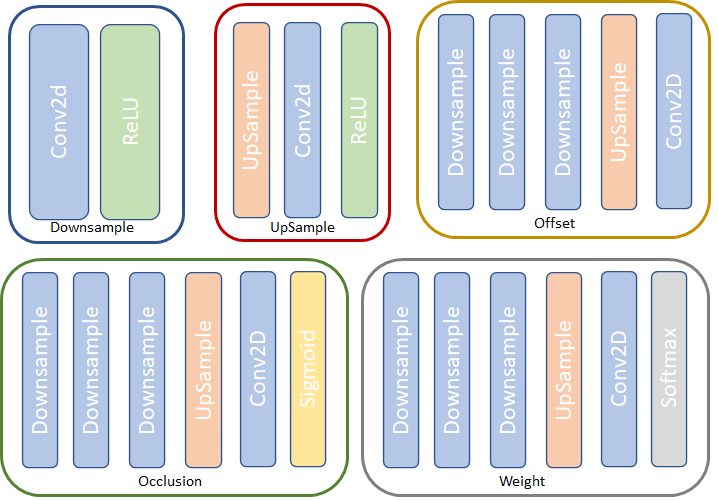}
    \caption{The figure shows the constituent layers of each block in the AdaCoF sub-network.}
    \label{fig:adacof_const}
\end{figure}
\begin{figure}[h]
    \centering
    \includegraphics[width=8cm]{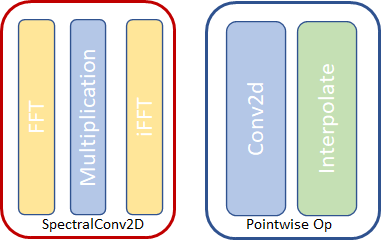}
    \caption{The figure shows the constituent operations performed in each of the blocks of the Neural Operator sub-network}
    \label{fig:nio_const}
\end{figure}
To overcome the issue of restricted kernel size in kernel-based methods, \cite{niklaus2017video} introduces adaptive separable convolution that uses deep convolutional neural networks and runs pairs of 1D kernels on the input frames to capture large motions. In addition to kernel-based methods, there are phase-based methods like \cite{meyer2015phase} that use per-pixel phase modification to compute a smooth interpolated frame without having to perform any flow estimation.

The above-discussed methods show exceptional results on well-known datasets like Vimeo-90K, but a major drawback is that these models take too long to converge. For example, DAIN \cite{bao2019depth} takes 5+ days to converge on an NVIDIA Titan X (Pascal) GPU for around 40 epochs and a batch size of 2; CAIN \cite{choi2020channel} is trained for 200 epochs with a batch size of 32 and takes around 4 days to converge on a single Titan Xp GPU.

\section{Proposed Approach}
AdaFNIO aims to overcome one of the common limitations exhibited by models that use convolutional layers - a variance to scale and makes the model resolution independent. Adaptive collaboration of flows, or AdaCoF, is a state-of-the-art architecture that improves the degrees of freedom of complex motions in the frames. The AdaCoF architecture offers a very generalized solution to determining the flows, irrespective of the number of pixels and their location in the frame \cite{lee2020adacof}. To generalize AdaCoF to any arbitrary resolution, we propose connecting neural operator layers to the model to capture finer information that is otherwise hard to generalize at higher resolutions. The final output is a weighted sum of the features learned by the neural operator and the flows learned by the base AdaCoF model. 

The proposed model AdaFNIO takes as input a couplet of frames and generates the intermediate frame. As shown in equation \ref{eq:1}, $I_0$ and $I_1$ are the input frames and $I_{0.5}$ is the interpolated frame. $w_1$ and $w_2$ are weights chosen for the features generated by the two models, which are tuned during training. If $I'_{0.5}$ is the ground truth and $\N$ is the neural operator then the resulting frame generation process is described as follows 
\begin{equation}
\label{eq:1}
    I_{0.5} = w_1 \N( I_0,I_1) + w_2 AdaCoF(I_0, I_1) 
\end{equation}
The interpolated frame quality is measured as PSNR( $I'_{0.5}$, $I_{0.5}$ ) and SSIM( $I'_{0.5}$, $I_{0.5}$ ).

\subsection{Operator Learning}

As proposed in \cite{li2020fourier}, the neural operator learns the mapping between two infinite dimensional spaces from a finite collection of input-output pairs. In our case, for frame interpolation, the infinite spaces are the functions corresponding to the trajectories of the constituent objects while the input-output pairs are the triplets representing micro-movement samples. 

Let $\A$ and $\U$ be the two such function spaces. Any frame $a$ is sampled from $\A$, and the corresponding output frame $u$ is sampled from $\U$. The neural operator $\G$ learns the mapping between $a$ and $u$ and is an approximation of the function mapping $\G$ from $\A$ to $\U$. The input image frames are treated as a collection of a continuous set of tokens, which are fed to the layers that perform token mixing to generate an embedding. The output features are generated from the low-dimensional embedding. 
The operator is a sequence of $\G_t$, for each layer $t \in \{1,2, ... N\}$,  each of which is a linear integral operator followed by non-linear activation. A kernel function is used within the integral operation and this becomes the basis of global convolution.  
The overall operation is given by equation \ref{eq:2}. 
\begin{equation}
\label{eq:2}
    \G_tv_t(x) = \sigma(\int \kappa_{t-1}(x,y)v_{t-1}d\mu_t(y) + W_tv_{t-1}(x))
\end{equation}
In the above equation, $\kappa_i$ is the kernel function with measure $\mu_i$, which along with the integral operator, performs a global linear operation. $W$ performs the point-wise linear operation. $\sigma$ is the non-linear activation. $v_t$ is the output of layer t. 
Iterative updates are performed as denoted by equation \ref{eq:3}
\begin{equation}
\label{eq:3}
    v_{t+1}(x) = \sigma(Wv_t(x) + (K_\phi v_t)(x))
\end{equation}
In the above equation, $\sigma$ is the non-linear activation and $W$ is the linear transformation and $K$ is the kernel operator parameterized by $\phi$.
The kernel operator is not explicitly represented in the spatial domain but is learned in the Fourier space directly through FFT and is represented as a periodic function R. The learned features are then added to the features learned by a point-wise operator through regular 2D convolution. This process is represented in equation \ref{eq:4}.
\begin{equation}
\label{eq:4}
    v_{t+1}(x) = \F^{-1}(R_\phi.\F(v_t))(x)
\end{equation}
In the above equation, R is the Fourier transform of a periodic function parameterized by $\phi$ and $\F$ denotes the Fourier Transform, which is given by equation \ref{eq:4}
\begin{equation}
\label{eq:4}
    (\F f)_j(k) = \int_D f_j(x)e^{-2i\pi \langle x,k \rangle}dx
\end{equation}
$\F^{-1}$ denotes the inverse Fourier Transform and is given by equation \ref{eq:5}
\begin{equation}
\label{eq:5}
    (\F^{-1}f)_j(x) = \int_D f_j(k)e^{2i\pi \langle x,k \rangle}dk    
\end{equation}
The above equations represent Fourier Transform in continuous space. However, images are discrete and therefore, Fast Fourier Transforms are applied.
The Fast Fourier transform in $d$ dimensional space is defined by equation \ref{eq:6}. 
\begin{equation}
\label{eq:6}
    (\hat{\F}f)_l(k) = \sum_{x_1=0}^{s_1-1} ... \sum_{x_d=0}^{s_d-1} f_l(x_1 ... x_d)e^{-2i\pi\Sigma_{j=1}^d \frac{x_jk_j}{s_j}}
\end{equation}
Inverse Fast Fourier Transform is given by equation \ref{eq:7}
\begin{equation}
\label{eq:7}
    (\hat{\F}^{-1}f)_l(x) = \sum_{k_1=0}^{s_1-1} ... \sum_{k_d=0}^{s_d-1} f_l(k_1 ... k_d)e^{2i\pi\Sigma_{j=1}^d \frac{x_jk_j}{s_j}} 
\end{equation}
The convolution operation in the Fourier space with the kernel tensor $R$, then becomes point-wise multiplication and is denoted by equation \ref{eq:9}.
\begin{equation}
\label{eq:9}
    (R (\F v_t))_{k,l} = \sum_{j=1}^{d_v} R_{k,l,j}(\F v_t)_{k,j}
\end{equation}

\subsection{Frames as a continuous set of tokens}
The input to the model is a set of frames, which are images. Neural operators have been traditionally applied to solve partial differential equations or PDEs where the input is the discretization of a continuous vector field. Images, on the other hand, have distinct objects, sharp edges and discontinuities. The frames can be broken down into a set of tokens, where each token is a patch within the frame. The model performs global convolution as a token mixing operation in the Fourier space. The tokens are extracted in the initial convolution layers with shared weights. The kernel size and the stride length determine the dimensions of the tokens. To account for non-periodicity in images, a local convolution layer is added after each global convolution. 

\subsection{Resolution Invariance and Quasi-linear Complexity}
The token features are learned in the Fourier space, which is invariant to the resolution of the input image. This was shown in \cite{guibas2021adaptive}. This allows the model to exhibit the property of zero-shot super-resolution; that is, the model can be trained on one resolution and tested on any arbitrary resolution. 

The multiplication is done on the lower $(k_x, k_y)$ Fourier modes, which is restricted to be at most $(n/2, m/2)$, where $(n, m)$ is the resolution of the image. If the weight matrix has a dimension of (p, p), then the time complexity of global convolution is $O(N log(N) p^2)$, where N is the length of the token sequence, as shown in \cite{guibas2021adaptive}.

The model contains two components - The interpolation layer, which performs a linear operation on the two frames and couples them together with a common weight matrix. The second component is the UNO network, as proposed in Rahman et al. \cite{rahman2022u}, which is a neural operator network. It performs convolution operations in the Fourier Space. Our model, NIO, initially contracts the input space to extract the key features of the image, which is equivalent to performing a dimensionality reduction. The encoder is followed by the decoder layers, which expand space back to its original size. The input takes two channels, with each channel corresponding to an input frame, while the output has a single channel corresponding to the output frame. 

\subsection{AdaFNIO Architecture}

The AdaFNIO model comprises of two sub-networks as shown in figure \ref{fig:overall_arch}. The AdaCoF network comprises of a U-Net implemented with convolutional layers and 6 networks that calculate 2 pairs of weight and offset vectors for the two frames. The individual blocks used in these networks are shown in figure \ref{fig:adacof_const}.

The neural operator network initially extracts the tokens through a sequence of convolution layers with shared weights. This means that the two frames are fed to the convolution layers recursively. This forms the input to the spectral convolution layers that perform global convolution in Fourier space. The pointwise operation layers perform local convolution and resize the frames to the required dimensions. The output features of each layer are a weighted sum of these two outputs. Through a sequence of spectral convolution and pointwise operator layers, the input tensor is downsampled and the latent embedding is generated. This embedding is used to generate the output frame through another sequence of spectral convolution and pointwise operator layers that upsample the input tensor successively. The constituent components of the spectral convolution block and pointwise operator block are shown in figure \ref{fig:nio_const}.

The spectral convolution layer only preserves the low-frequency Fourier models and ignores the high-frequency modes, which are too specific to the particular input and if these modes are learned, they overfit to the input as shown in Li et al.\cite{li2020fourier}. After applying the weights, the tensors are projected back into the spatial domain and non-linear activation is applied to recover the high-frequency points.

\begin{figure*}[h]
    \centering
    \includegraphics[width=1\textwidth]{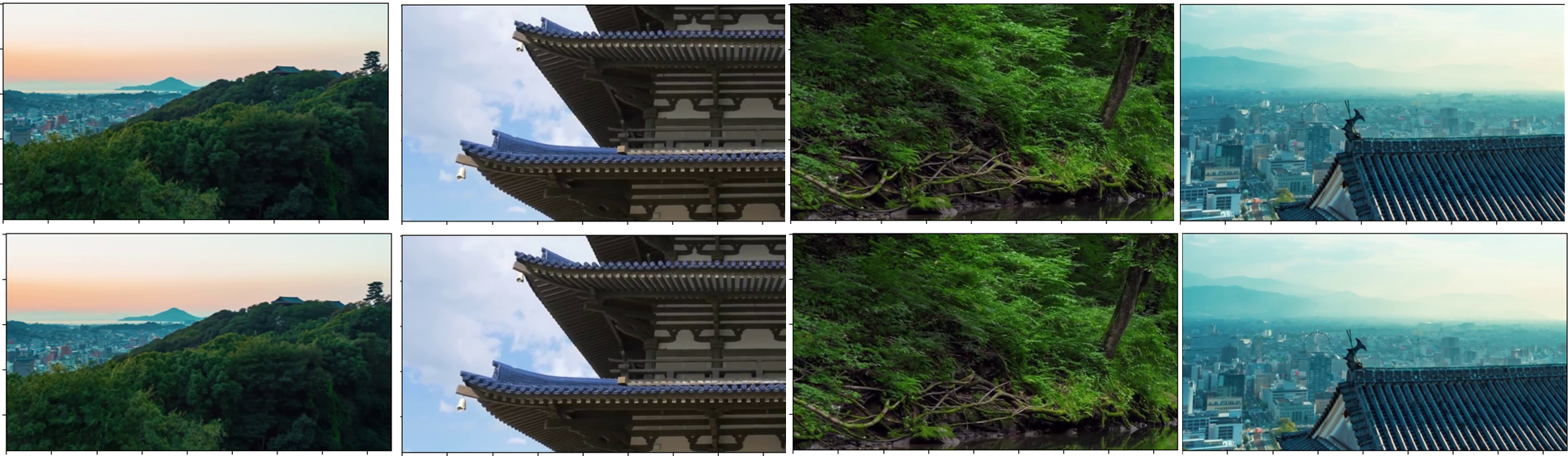}
    \caption{This figure is a visual comparison of results generated by AdaFNIO against high-resolution (1080p) stock footage of the Japanese landscape. While the visual differences are hard to discern, the AdaFNIO has a slightly better quantitative performance against AdaCoF. The top row is the output generated by AdaFNIO and the bottom row is the ground truth.}
    \label{fig:diffscenes}
\end{figure*}

\subsubsection{Initial convolution layers}
The first two layers of the networks are local convolution layers which apply the linear operator to the individual frames recursively to extract the tokens. This layer captures and aligns the common features of the two images together. 
\subsubsection{Spectral convolution and pointwise operator layers}

The downsampler block has 4 contraction layers, which reduce the input size from 256x256 to 32x32. The number of Fourier modes retained in each layer are 42, 21, 10, and 5,  respectively. The upsampler block is symmetric to the downsampler and expands from 32x32 to 256x256, with 5, 10, 21, and 42 Fourier modes, respectively. As opposed to using regular convolution to perform downsampling and upsampling, our NIO model uses spectral convolution to perform feature extraction and dimensionality reduction in Fourier space which is followed by %THIS INFORMATION SEEMS A BIT LOW-LEVEL FOR AN OVERVIEW, MAYBE MORE IT SOMEWHERE WHERE IT'S MORE RELEVANT. DO YOU HAVE AN OVERVIEW AND ARCHITECTURE DIAGRAM? 
GELU activation function to recover high-frequency information. 
\subsubsection{Loss Functions}
The model uses two loss functions. The L1 loss function is used to initially train the model. This loss is given by equation \ref{eq:10}
\begin{equation}
    \label{eq:10}
    L_1 = || I_{AdaFNIO} - I_{GT} ||_1
\end{equation}
The other loss function that is used is perceptual loss, which is used for fine-tuning the model. The loss is generated by the feature extractor of the pre-trained VGG22 neural network. This loss is given by equation \ref{eq:11}
\begin{equation}
    \label{eq:11}
    L_{vgg} = ||F(I_{AdaFNIO}) - F(I_{GT})||
\end{equation}
The overall loss function used during fine-tuning is a combination of L1 and VGG loss functions, with higher weights given to the L1 loss function. This is denoted by equation \ref{eq:12}
\begin{equation}
    \label{eq:12}
    L = L_1 + 0.01 * L_{vgg}
\end{equation}
\begin{figure*}[h]
    \centering
    \includegraphics[width=1\textwidth]{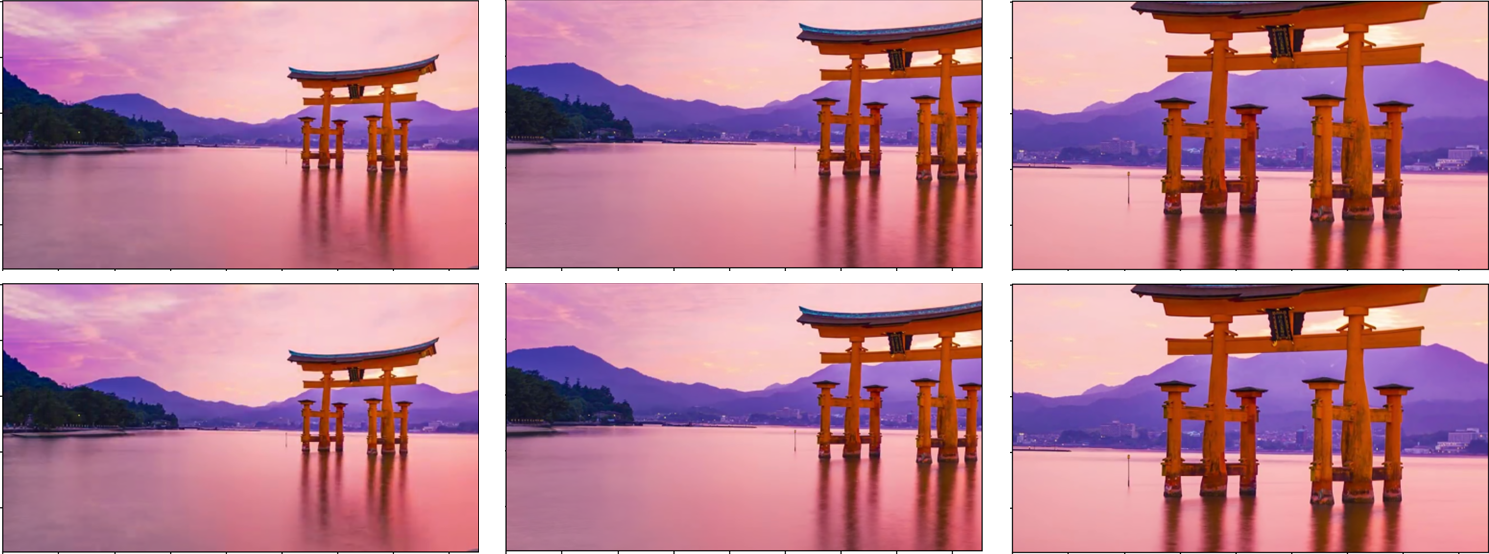}
    \caption{The above figure highlights the resolution invariance property exhibited by the architecture. The models were trained on 256x256 patches of the Vimeo90K dataset but tested against high-resolution stock footage of the Japanese landscape. In this figure, the top row is the generated output and the bottom row is the ground truth. The left row is of 480p resolution, the center one 720p and the right one 1080p}
    \label{fig:resolution_invariance}
\end{figure*}

\subsection{Frame Generation}
The intermediate frame is generated after applying a sequence of spectral convolution layers that perform global convolution in Fourier space. Let the input frames be represented as $I_i$ and $I_j$ and the initial weight matrix be represented as $V_f$. Let the initial convolution layers be represented with $C_t$. Let the downsampler layer be represented as $E$ and the upsampler layer be represented as $D$. Let the weights be represented as $W$ and bias be represented with $B$ and AdaCoF network be represented by $Ada$.  The pipeline is as follows
\begin{equation}
    C_t(I_i, I_j) = V_f * I_i + V_f * I_j + B
\end{equation}

\begin{equation}
    I_{0.5} = W_{NIO} * D(E(C_t(I_i, I_j))) + Ada(I_i, I_j)
\end{equation}
\subsection{Training}
For the training process, Vimeo90K triplet dataset was used. The dataset has 73,171 3-frame sequences, of which 58,536 frames were used for training and the remaining 14635 were used for validation. The L1 loss was used for the first 80 epochs. For the finetuning process, 11,000 random frames were used to tune the model for another 20 epochs with perceptual loss. The model was tested against 1080p stock footage of a Japanese landscape taken from YouTube channel 8K World. This video was chosen because the footage was shot in very high resolution. The model was trained for 100 epochs on Nvidia A100 GPU. The frames were randomly cropped to 256x256 patches. However, the frames were not randomly flipped, scaled, or rotated in order to test the invariance properties of the architecture.
\subsubsection{Dataset}
The AdaFNIO model was validated on these three frequently used benchmark datasets (Vimeo90K, DAVIS, UCF101) as well as one specialty dataset (DISFA+), which focuses on videos with human faces. 
\begin{table*}[t]
\centering
\begin{tabular}{l l l l l l l l l l l l}
\hline
Model & Parameters (M) & Epochs & \multicolumn{2}{l}{\textit{\textbf{Vimeo90K}}} & \multicolumn{2}{l}{\textit{\textbf{DAVIS}}} & \multicolumn{2}{l}{\textit{\textbf{UCF101}}} & \multicolumn{2}{l}{\textit{\textbf{DISFA+}}}\\
      &      &      & PSNR          & SSIM         & PSNR        & SSIM        &       PSNR       &    SSIM     & PSNR  & SSIM\\
      \hline
ToFlow \cite{xue2019video}     &    1.4        & -&   33.73            &  0.968            &  - & - & 34.58          &     0.966   &-&-    \\
IFRNET-S \cite{kong2022ifrnet}     &    2.8       & 300 &  35.59             &     0.978         &  - & - &   35.28        &    0.969 & 38.85 & 0.961         \\

VFIT-S  \cite{shi2022video}    &    7.5        &  100             &     36.48         &    0.976         &    \textbf{27.92} & 0.885   & \textbf{33.36} & \textbf{0.971} & 39.25 & 0.964     \\
SoftSplat \cite{niklaus2020softmax}     &   7.7     &  -  &    35.76        & 0.972                          &     27.42        &    0.878   & 35.39 & 0.952  & 38.33 & 0.954    \\
RIFE  \cite{huang2020rife}    &    9.8        &  25   &    35.62      &   0.978           & - & - &    35.28 & 0.969        &   38.84  & 0.961       \\
BMBC \cite{park2020bmbc}     &  11.0     &  -   &      34.76      & 0.965                      &      26.42       &    0.868  &  35.15 & 0.968 & - & -      \\
ABME  \cite{park2021asymmetric}    &    18.1        & -   & 36.18          &   \textbf{0.980}           & - & - & 35.38 & 0.969            &      -&-      \\
SepConv \cite{zhang2018video}     &  21.6     &  -   &      33.60       & 0.944                            &   26.21          &    0.857 & 34.78 & 0.966 & 38.70 & 0.959         \\
AdaCof \cite{lee2020adacof}     &    21.8        &  50             &  34.47            &     0.973        &   26.49 & 0.866   &  34.90 & 0.968 & 38.98 & 0.961       \\
DAIN  \cite{bao2019depth}    &   24.0      &  40 &        33.35        & 0.945                         &   26.12          &    0.870   & 34.99 & 0.968 & 35.00 & 0.956       \\
QVI \cite{liu2020enhanced}     &    29.2        &   200   &    35.15     &  0.971            &     27.17 & 0.874        &      32.89 & 0.970 & - & -        \\
SuperSloMo \cite{jiang2018super}     &  39.6    &   500   &   32.90       &  0.957                           &    25.65         &    0.857  & 32.33 & 0.960 & - & -       \\
FLAVR \cite{kalluri2020flavr}     &    42.4    &   200 &     36.30          & 0.975              &      27.44       &  0.874    & 33.33 & 0.971 & - & -       \\
CAIN \cite{choi2020channel}     &   42.8     & 200   &        34.76        & 0.970                       &    27.21         &     0.873  & 34.91 & 0.969 & - & -      \\
\textbf{AdaFNIO}      &    88.9       &   100           & \textbf{36.50} &      0.976        &      27.90       &  \textbf{0.888}    & 34.88 & 0.970 & \textbf{39.30} & \textbf{0.965}      \\
\end{tabular}
\caption{The table showing the quantitative performance on Vimeo90K, DAVIS, UCF101 and DISFA+ dataset. The AdaFNIO model is compared against the quantitative performance of other models, as presented in Kong et al. \cite{kong2022ifrnet} and Shi et al. \cite{shi2022video} }
\label{tab:main_res}
\end{table*}

\begin{itemize}
    \item \textbf{Vimeo90K Dataset} \cite{xue2019video} The Vimeo90K dataset is built from 89,800 clips taken from the video streaming site Vimeo. It contains a large variety of scenes. For this project, the triplet dataset was used. The dataset contains 73,171 3-frame sequences, all of which have a resolution of 448x256. 
    \item \textbf{Davis-90} The modified Densely Annotated Video Segmentation dataset contains frames from various scenes.  These frames are partitioned into triplet sets and used for testing the performance of the model. 
    \item \textbf{UCF101 Dataset} The preprocessed UCF101 dataset is a collection of scenes that have been partitioned into triplets. This dataset is also used for testing the models. 
    \item \textbf{DISFA+ Dataset} The DISFA+ or the Denver Intensity of Spontaneous Facial Action Database consists of a large set of facial expression sequences, both posed and non-posed. The dataset has multiple subjects of different ethnicities exhibiting various facial expressions and is a comprehensive dataset to study micro facial expressions. This dataset was chosen due to the increase in the prevalence of video meetings and social media videos, many of which predominantly features human faces. 
\end{itemize}

The DISFA+ dataset was processed into triplets and the model was trained to predict the second frame from the first and third frames. The Vimeo90K dataset was used to provide a comparison benchmark against other deep learning-based interpolation approaches, while the DISFA+ dataset was used to predict facial expressions from up close. This served as a test to determine the ability of neural operator-based models to interpolate minute facial muscle movements.
The frames were resized to 256x256 due to memory and GPU constraints. 
\begin{figure}[h]
    \centering
    \includegraphics[width=8cm]{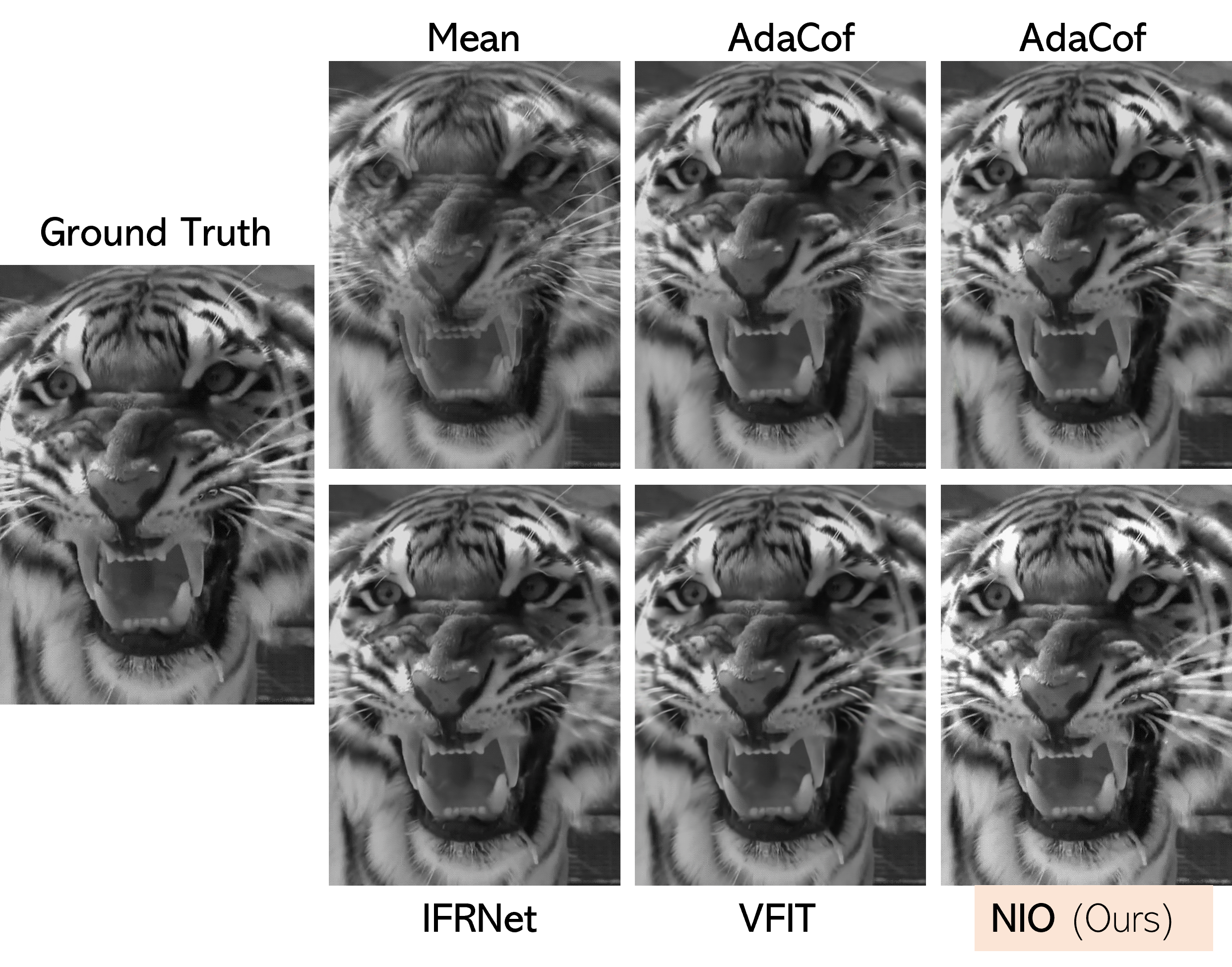}
    \caption{This figure provides a visual comparison of results across various models. NIO achieves visually comparable results with other models.}
    \label{fig:diffmodels}
\end{figure}
\subsubsection{Hyperparameters}
The models were built using Pytorch and trained on Nvidia A100 GPUs. The Fourier modes used for the layers are 5, 10, 21 and 42. The batch size was set to 32 and the learning rate was set to 0.0001. The weight for the NIO base model was set to 0.01. The training was done using Cuda 11.0. Adam optimizer was used with a weight decay of 0.0001, $\beta_1$ of 0.9 and $\beta_2$ of 0.999. The loss function used for training was a mean squared error (MSE) or L2 Loss. 
\section{Experiments}

\begin{figure}[h!]
    \centering
    \includegraphics[width=8cm]{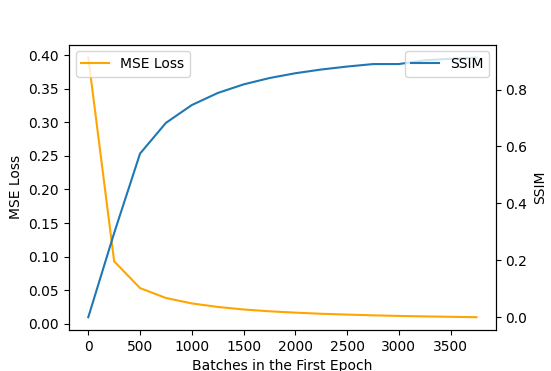}
    \caption{Loss trend in the \textbf{First Epoch} for every 250th batch of low-resolution Vimeo90K Dataset on the pure NIO base model.}
    \label{fig:loss_pure_NIO}
\end{figure}
\subsection{Evaluation}
\subsubsection{Quantitative performance}
In this section, we provide a comparison benchmark against the other state-of-the-art models with SSIM and PSNR as the evaluation metrics. Table \ref{tab:main_res} shows the quantitative performance against other state-of-the-art models. AdaFNIO has the best PSNR (36.50) on the Vimeo90K dataset, and the best SSIM (0.888) on the DAVIS dataset and outperforms every other model against DISFA+ dataset. 
The reason for better performance against DISFA+ dataset is that neural operator-based models perform well on periodic images with smooth edges. Talking head videos have the fewest number of objects and fewer edges within the frames and thus, neural operators outperform in that situation. This phenomenon was first identified by \cite{guibas2021adaptive} and empirically verified by us. 

\subsubsection{Qualitative performance}
Qualitatively, the images generated by the AdaFNIO model are indistinguishable from the ground truth images. However, we performed experiments against AdaCoF model to test the extent of generalization at higher resolutions. AdaFNIO had better SSIM and PSNR values at higher-resolution videos. Figure \ref{fig:diffscenes} depicts the results generated by AdaFNIO against Japanese landscape stock footage video. These frames were captured at 1080p resolution. \\

Figure \ref{fig:resolution_invariance} shows the outputs generated by the model at three different resolutions 480p, 720p and 1080p, respectively, and there is no performance degradation as the resolution increases. Figure \ref{fig:diffmodels} highlights the performance of AdaFNIO against other state-of-the-art models. 

\subsubsection{Comparisons against baseline AdaCoF model}
In this section, we show quantitative differences between the frames generated by AdaCoF model and AdaFNIO model in two settings - varying resolutions and varying frame rates. The models were tested against the Japanese stock footage video at 30fps, with 9,894 frames.\\
\textbf{Varying resolutions} At lower resolution, we observed that AdaFNIO and AdaCoF had similar performance but as the resolution increased, AdaFNIO performed slightly better than the AdaCoF model. These SSIM values at different resolutions are shown in table \ref{tab:diff_res}
\begin{table}[h]
    \centering
    
    \begin{tabular}{c c c c}
    \hline
      \textbf{Model}   &  \textbf{480p} & \textbf{720p} & \textbf{1080p}\\
      \hline
      AdaFNIO &  98.276 & 98.792 & 99.207\\
      AdaCoF &   98.276 & 98.790 & 99.199\\
    \end{tabular}
    
    \caption{The SSIM values against Japanese stock footage video captured at different resolutions}
    \label{tab:diff_res}
\end{table}

\textbf{Varying frame rates} To test the performance of the model with missing frames, the model was evaluated in three settings against 480p resolution frames - when every alternate frame was dropped, when 2 consecutive frames were dropped and when 4 consecutive frames were dropped. AdaFNIO slightly outperformed AdaCoF in all three settings and the SSIM values are shown in table \ref{tab:diff_drop}

\begin{table}[h]
    \centering
    
    \begin{tabular}{c c c c}
    \hline
      \textbf{Model}   &  \textbf{drop 1} & \textbf{drop 2} & \textbf{drop 4}\\
      \hline
      AdaFNIO &  98.257 & 94.909 & 89.988\\
      AdaCoF &   98.256 & 94.899 & 89.969\\
    \end{tabular}
    
    \caption{The SSIM values against Japanese stock footage video captured at fixed resolution of 480p but with varying frame rates}
    \label{tab:diff_drop}
\end{table}

\subsection{Ablation Study}
To perform the ablation study, we tested different neural operator models against variants of Vimeo90K dataset in different settings. Pure neural operator models are highly sensitive to sharp edges and generally perform well on low-resolution frames due to them having relatively smooth boundaries and fewer sharp objects. To remedy this issue, the neural operator was combined with AdaCoF in the final version of AdaFNIO. We shall discuss our findings on pure neural operator models in the following subsections. 
\begin{figure}
    \centering
    \includegraphics[width=8cm]{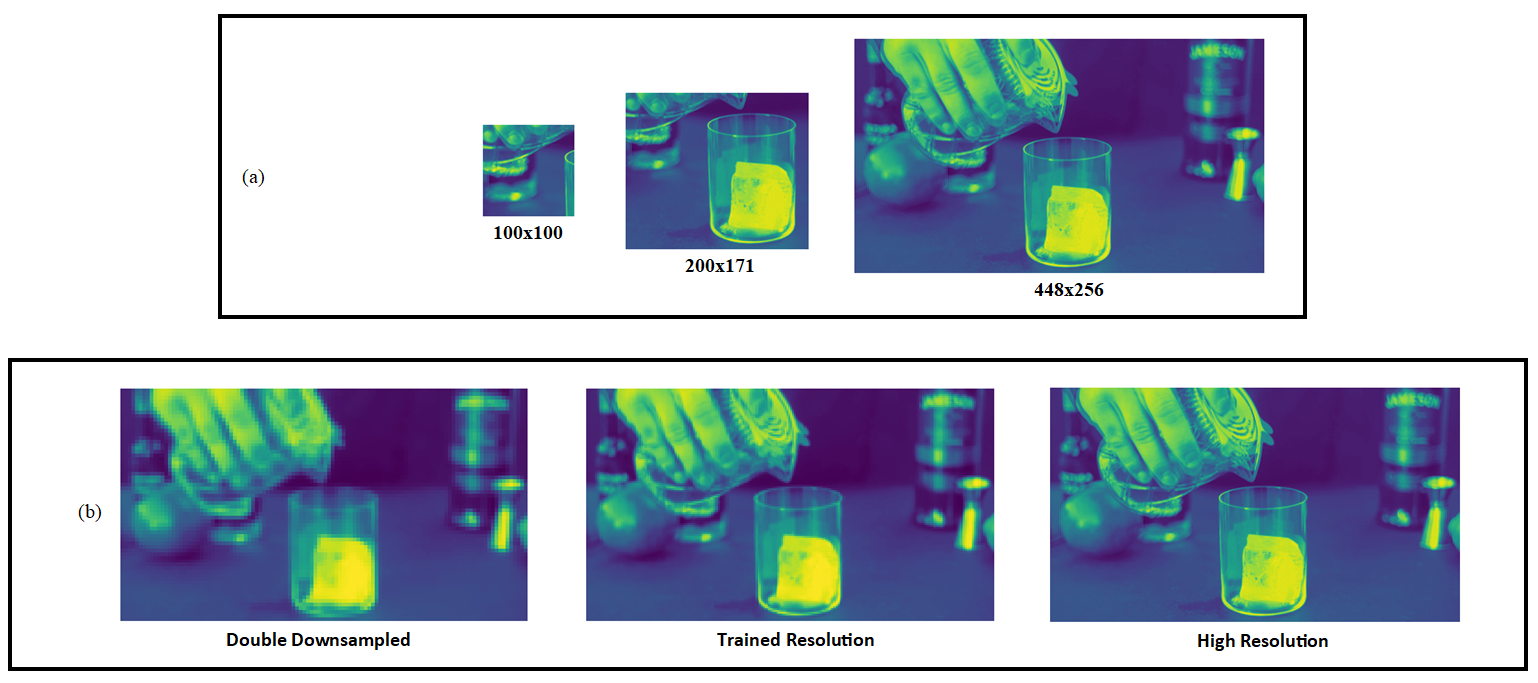}
    \caption{The figure illustrates the resolution invariance of the pure NIO Model without the AdaCoF layers. The model was trained on downsampled low-resolution 100x100 images, the same resolution as the leftmost image in (a). The other two images in (a) are 200x171 and 448x256, respectively. The network was \textbf{never} previously trained on images of those sizes. In row (b), the image on the right is the high-resolution image predicted by the model. The image in the middle is the downsampled image, whose resolution is comparable to the training dataset for the model, while the image on the left has been downsampled twice.}
    \label{fig:my_label}
\end{figure}

 \begin{figure}[h]
    \centering
    \includegraphics[width=8cm]{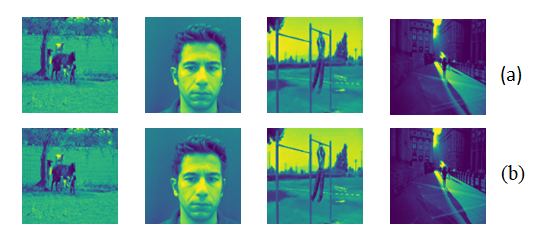}
    \caption{Visual Comparison of Ground Truth and Generated images in the pure NIO model against low-resolution Davis-90, DISFA+, UCF101 and Vimeo90K. Row (a) represents the generated images and row (b) represents the ground truth.}
    \label{fig:my_label}
\end{figure}
We built two models - a basic NIO model with a single NIO block with 3.42 million parameters and a lighter model - NIO-Small (NIO-S), with 1.18 million parameters and trained the model on the grayscale Vimeo-90K dataset. In the NIO-S model, the decoder is replaced with regular convolution2D and the encoder retains the channels instead of downsampling. The NIO-S model was not as accurate as the NIO model, but it was faster and took half as much time per epoch. On the A30 GPU, the NIO model took 5 and a half minutes per epoch, while the NIO-S model took 3 minutes and 39 seconds per epoch on average. This model only had three spectral convolution layers. The two models approached an SSIM of 0.90 within the first epoch, as shown in figure \ref{fig:loss_pure_NIO}. Table \ref{tab:NIO_comps} highlights the differences between a single NIO block but with different number of spectral layers when tested against low resolution (50x50) DISFA+ dataset. 
\begin{table}[t]
\centering
\begin{tabular}{l l l l l l }
\hline
Model & Spectral & Time/Epoch  & \multicolumn{2}{l}{DISFA+}\\
      & Layers     & fixed batch  & PSNR          & SSIM \\
      \hline
NIO-S      &    4        & 00:03:39  &  39.25& 0.987 \\
NIO      &    8        &   00:05:43 & 38.84 &0.954      \\
\end{tabular}
\caption{The table showing the differences between NIO-S and NIO models }
\label{tab:NIO_comps}
\end{table}

\textbf{Neural Operators and low-resolution RGB images}
Three models were tested against low-resolution RGB version of Vimeo90K dataset at 2 different resolutions of 100x100 and 85x85. These models were all trained for 50 epochs at a batch size of 32. The first model was the NIO fine model with 8 NIO blocks connected to form a residual network, the second was an NIO base model with 4 NIO blocks, and lastly, the AFNO Net with channel attention, tested against two patch sizes of 1x1 and 2x2. The models were tested against both normalized and non-normalized datasets. The SSIM values after convergence are shown in table \ref{tab:no_comp}.

\begin{table}[h]
    \centering
    
    \begin{tabular}{c c c c c}
    \hline
      \textbf{Model}   &  \textbf{100x100} & \textbf{100x100 N} & \textbf{85x85} & \textbf{85x85 N}\\
      \hline
      NIO base &  84.526 & 86.438 & 90.235 & 91.876\\
      NIO fine &   88.781 & 89.623 & 92.693 & 93.451\\

      AFNO 1x1 &   74.376 & 74.298 & 76.851 & 76.394\\
      AFNO 2x2 &   77.368 & 77.925 & 79.422 & 79.631\\
    \end{tabular}
    
    \caption{The SSIM values low resolution RGB Vimeo90K dataset for pure neural operator models. The N refers to normalized dataset}
    \label{tab:no_comp}
\end{table}

It can be observed from table \ref{tab:no_comp} that AFNO network, which uses channel attention, performed poorly in every setting. NIO models that used downsamplers and upsamplers were quick to converge, as seen in table \ref{tab:NIO_comps} and performed better than AFNOs. Therefore, the NIO models were used as the neural operator component in the final AdaFNIO model. 

\subsubsection{Normalization}
Min-Max normalization was applied to the dataset as a preprocessing step, reducing the range of values that a pixel can take. The models performed slightly worse on the Vimeo90K dataset when the input was normalized. However, these models performed better on the other datasets after being trained on the normalized grayscale 100x100 Vimeo90K dataset, thus showing that normalization as a preprocessing step helps the models generalize better on data they have not seen before. 
\begin{table}[h]
\centering
\begin{tabular}{l l l l l l }
\hline
Model   & \multicolumn{2}{l}{UCF101}\\
       &  PSNR          & SSIM \\
      \hline
NIO      &  36.54 & 0.970 \\
NIO + Norm     & 36.84 &0.954      \\
\end{tabular}
\caption{The table showing the differences between Normalizing the input and not normalizing the input }
\label{tab:my_label}
\end{table}
\section{Conclusion}
In this paper, we have presented a neural operator-based architecture for performing frame interpolation. The model is powerful, resolution invariant, and discretization invariant, and achieves state-of-the-art performance on unseen datasets.
The model has proven effective in capturing information in tiny regions of the image (tokens) and generalizing well in larger images.
The AdaFNIO model has only been trained on a triplet dataset with consecutive frames. However, it remains to be seen how well it performs when trained against larger sequences of frames. 
Secondly, the resolution invariance is an important property of the neural operator and it remains to be seen whether this can be used to improve the resolution of the images.
{\small
\bibliographystyle{ieee_fullname}
\bibliography{egpaper_final}

\begin{thebibliography}{10}\itemsep=-1pt

\bibitem{ali2021deep}
Sharib Ali, Felix Zhou, Adam Bailey, Barbara Braden, James~E East, Xin Lu, and
  Jens Rittscher.
\newblock A deep learning framework for quality assessment and restoration in
  video endoscopy.
\newblock {\em Medical image analysis}, 68:101900, 2021.

\bibitem{bao2019depth}
Wenbo Bao, Wei-Sheng Lai, Chao Ma, Xiaoyun Zhang, Zhiyong Gao, and Ming-Hsuan
  Yang.
\newblock Depth-aware video frame interpolation.
\newblock In {\em Proceedings of the IEEE/CVF Conference on Computer Vision and
  Pattern Recognition}, pages 3703--3712, 2019.

\bibitem{bharadwaj2021video}
Ashwin~R Bharadwaj, Hardik Gourisaria, and Hrishikesh Viswanath.
\newblock Video frame rate doubling using generative adversarial networks.
\newblock In {\em Computer Communication, Networking and IoT}, pages 463--474.
  Springer, 2021.

\bibitem{Caelles_arXiv_2019}
Sergi Caelles, Jordi Pont-Tuset, Federico Perazzi, Alberto Montes,
  Kevis-Kokitsi Maninis, and Luc {Van Gool}.
\newblock The 2019 davis challenge on vos: Unsupervised multi-object
  segmentation.
\newblock {\em arXiv:1905.00737}, 2019.

\bibitem{chen2017long}
Xiongtao Chen, Wenmin Wang, and Jinzhuo Wang.
\newblock Long-term video interpolation with bidirectional predictive network.
\newblock In {\em 2017 IEEE Visual Communications and Image Processing (VCIP)},
  pages 1--4. IEEE, 2017.

\bibitem{cheng2021audio}
Harry Cheng, Yangyang Guo, Jianhua Yin, Haonan Chen, Jiafang Wang, and Liqiang
  Nie.
\newblock Audio-driven talking video frame restoration.
\newblock {\em IEEE Transactions on Multimedia}, 2021.

\bibitem{cheng2020video}
Xianhang Cheng and Zhenzhong Chen.
\newblock Video frame interpolation via deformable separable convolution.
\newblock In {\em Proceedings of the AAAI Conference on Artificial
  Intelligence}, volume~34, pages 10607--10614, 2020.

\bibitem{choi2020channel}
Myungsub Choi, Heewon Kim, Bohyung Han, Ning Xu, and Kyoung~Mu Lee.
\newblock Channel attention is all you need for video frame interpolation.
\newblock In {\em Proceedings of the AAAI Conference on Artificial
  Intelligence}, volume~34, pages 10663--10671, 2020.

\bibitem{danier2022st}
Duolikun Danier, Fan Zhang, and David Bull.
\newblock St-mfnet: A spatio-temporal multi-flow network for frame
  interpolation.
\newblock In {\em Proceedings of the IEEE/CVF Conference on Computer Vision and
  Pattern Recognition}, pages 3521--3531, 2022.

\bibitem{guibas2021adaptive}
John Guibas, Morteza Mardani, Zongyi Li, Andrew Tao, Anima Anandkumar, and
  Bryan Catanzaro.
\newblock Adaptive fourier neural operators: Efficient token mixers for
  transformers.
\newblock {\em arXiv preprint arXiv:2111.13587}, 2021.

\bibitem{he2020video}
Gang He, Chang Wu, Lei Li, Jinjia Zhou, Xianglin Wang, Yunfei Zheng, Bing Yu,
  and Weiying Xie.
\newblock A video compression framework using an overfitted restoration neural
  network.
\newblock In {\em Proceedings of the IEEE/CVF Conference on Computer Vision and
  Pattern Recognition Workshops}, pages 148--149, 2020.

\bibitem{hu2021capturing}
Mengshun Hu, Jing Xiao, Liang Liao, Zheng Wang, Chia-Wen Lin, Mi Wang, and
  Shin’ichi Satoh.
\newblock Capturing small, fast-moving objects: Frame interpolation via
  recurrent motion enhancement.
\newblock {\em IEEE Transactions on Circuits and Systems for Video Technology},
  2021.

\bibitem{hu2022many}
Ping Hu, Simon Niklaus, Stan Sclaroff, and Kate Saenko.
\newblock Many-to-many splatting for efficient video frame interpolation.
\newblock In {\em Proceedings of the IEEE/CVF Conference on Computer Vision and
  Pattern Recognition}, pages 3553--3562, 2022.

\bibitem{huang2020rife}
Zhewei Huang, Tianyuan Zhang, Wen Heng, Boxin Shi, and Shuchang Zhou.
\newblock Rife: Real-time intermediate flow estimation for video frame
  interpolation.
\newblock {\em arXiv preprint arXiv:2011.06294}, 2020.

\bibitem{huang2022real}
Zhewei Huang, Tianyuan Zhang, Wen Heng, Boxin Shi, and Shuchang Zhou.
\newblock Real-time intermediate flow estimation for video frame interpolation.
\newblock In {\em European Conference on Computer Vision}, pages 624--642.
  Springer, 2022.

\bibitem{jiang2018super}
Huaizu Jiang, Deqing Sun, Varun Jampani, Ming-Hsuan Yang, Erik Learned-Miller,
  and Jan Kautz.
\newblock Super slomo: High quality estimation of multiple intermediate frames
  for video interpolation.
\newblock In {\em Proceedings of the IEEE conference on computer vision and
  pattern recognition}, pages 9000--9008, 2018.

\bibitem{kalluri2020flavr}
Tarun Kalluri, Deepak Pathak, Manmohan Chandraker, and Du Tran.
\newblock Flavr: Flow-agnostic video representations for fast frame
  interpolation.
\newblock {\em arXiv preprint arXiv:2012.08512}, 2020.

\bibitem{karargyris2010three}
Alexandros Karargyris and Nikolaos Bourbakis.
\newblock Three-dimensional reconstruction of the digestive wall in capsule
  endoscopy videos using elastic video interpolation.
\newblock {\em IEEE transactions on Medical Imaging}, 30(4):957--971, 2010.

\bibitem{kim2022cross}
Hannah~Halin Kim, Shuzhi Yu, Shuai Yuan, and Carlo Tomasi.
\newblock Cross-attention transformer for video interpolation.
\newblock {\em arXiv preprint arXiv:2207.04132}, 2022.

\bibitem{kim2018spatio}
Tae~Hyun Kim, Mehdi~SM Sajjadi, Michael Hirsch, and Bernhard Scholkopf.
\newblock Spatio-temporal transformer network for video restoration.
\newblock In {\em Proceedings of the European Conference on Computer Vision
  (ECCV)}, pages 106--122, 2018.

\bibitem{kong2022ifrnet}
Lingtong Kong, Boyuan Jiang, Donghao Luo, Wenqing Chu, Xiaoming Huang, Ying
  Tai, Chengjie Wang, and Jie Yang.
\newblock Ifrnet: Intermediate feature refine network for efficient frame
  interpolation.
\newblock In {\em Proceedings of the IEEE/CVF Conference on Computer Vision and
  Pattern Recognition}, pages 1969--1978, 2022.

\bibitem{kovachki2021neural}
Nikola Kovachki, Zongyi Li, Burigede Liu, Kamyar Azizzadenesheli, Kaushik
  Bhattacharya, Andrew Stuart, and Anima Anandkumar.
\newblock Neural operator: Learning maps between function spaces.
\newblock {\em arXiv preprint arXiv:2108.08481}, 2021.

\bibitem{krishnamurthy1999frame}
Ravi Krishnamurthy, John~W Woods, and Pierre Moulin.
\newblock Frame interpolation and bidirectional prediction of video using
  compactly encoded optical-flow fields and label fields.
\newblock {\em IEEE transactions on circuits and systems for video technology},
  9(5):713--726, 1999.

\bibitem{lee2020adacof}
Hyeongmin Lee, Taeoh Kim, Tae-young Chung, Daehyun Pak, Yuseok Ban, and
  Sangyoun Lee.
\newblock Adacof: Adaptive collaboration of flows for video frame
  interpolation.
\newblock In {\em Proceedings of the IEEE/CVF Conference on Computer Vision and
  Pattern Recognition}, pages 5316--5325, 2020.

\bibitem{li2020fourier}
Zongyi Li, Nikola Kovachki, Kamyar Azizzadenesheli, Burigede Liu, Kaushik
  Bhattacharya, Andrew Stuart, and Anima Anandkumar.
\newblock Fourier neural operator for parametric partial differential
  equations.
\newblock {\em arXiv preprint arXiv:2010.08895}, 2020.

\bibitem{liang2022vrt}
Jingyun Liang, Jiezhang Cao, Yuchen Fan, Kai Zhang, Rakesh Ranjan, Yawei Li,
  Radu Timofte, and Luc Van~Gool.
\newblock Vrt: A video restoration transformer.
\newblock {\em arXiv preprint arXiv:2201.12288}, 2022.

\bibitem{liu2020video}
Xiaozhang Liu, Hui Liu, and Yuxiu Lin.
\newblock Video frame interpolation via optical flow estimation with image
  inpainting.
\newblock {\em International Journal of Intelligent Systems},
  35(12):2087--2102, 2020.

\bibitem{liu2020enhanced}
Yihao Liu, Liangbin Xie, Li Siyao, Wenxiu Sun, Yu Qiao, and Chao Dong.
\newblock Enhanced quadratic video interpolation.
\newblock In {\em European Conference on Computer Vision}, pages 41--56.
  Springer, 2020.

\bibitem{liu2019deep}
Yu-Lun Liu, Yi-Tung Liao, Yen-Yu Lin, and Yung-Yu Chuang.
\newblock Deep video frame interpolation using cyclic frame generation.
\newblock In {\em Proceedings of the AAAI Conference on Artificial
  Intelligence}, volume~33, pages 8794--8802, 2019.

\bibitem{liu2017voxelflow}
Ziwei Liu, Raymond~A Yeh, Xiaoou Tang, Yiming Liu, and Aseem Agarwala.
\newblock Video frame synthesis using deep voxel flow.
\newblock In {\em Proceedings of the IEEE international conference on computer
  vision}, pages 4463--4471, 2017.

\bibitem{luo2022bi}
Yao Luo, Jinshan Pan, and Jinhui Tang.
\newblock Bi-directional pseudo-three-dimensional network for video frame
  interpolation.
\newblock {\em IEEE Transactions on Image Processing}, 2022.

\bibitem{mavadati2016extended}
Mohammad Mavadati, Peyten Sanger, and Mohammad~H Mahoor.
\newblock Extended disfa dataset: Investigating posed and spontaneous facial
  expressions.
\newblock In {\em proceedings of the IEEE conference on computer vision and
  pattern recognition workshops}, pages 1--8, 2016.

\bibitem{meyer2015phase}
Simone Meyer, Oliver Wang, Henning Zimmer, Max Grosse, and Alexander
  Sorkine-Hornung.
\newblock Phase-based frame interpolation for video.
\newblock In {\em Proceedings of the IEEE conference on computer vision and
  pattern recognition}, pages 1410--1418, 2015.

\bibitem{niklaus2018context}
Simon Niklaus and Feng Liu.
\newblock Context-aware synthesis for video frame interpolation.
\newblock In {\em Proceedings of the IEEE conference on computer vision and
  pattern recognition}, pages 1701--1710, 2018.

\bibitem{niklaus2020softmax}
Simon Niklaus and Feng Liu.
\newblock Softmax splatting for video frame interpolation.
\newblock In {\em Proceedings of the IEEE/CVF Conference on Computer Vision and
  Pattern Recognition}, pages 5437--5446, 2020.

\bibitem{niklaus2017video}
Simon Niklaus, Long Mai, and Feng Liu.
\newblock Video frame interpolation via adaptive separable convolution.
\newblock In {\em Proceedings of the IEEE International Conference on Computer
  Vision}, pages 261--270, 2017.

\bibitem{park2020bmbc}
Junheum Park, Keunsoo Ko, Chul Lee, and Chang-Su Kim.
\newblock Bmbc: Bilateral motion estimation with bilateral cost volume for
  video interpolation.
\newblock In {\em European Conference on Computer Vision}, pages 109--125.
  Springer, 2020.

\bibitem{park2021asymmetric}
Junheum Park, Chul Lee, and Chang-Su Kim.
\newblock Asymmetric bilateral motion estimation for video frame interpolation.
\newblock In {\em Proceedings of the IEEE/CVF International Conference on
  Computer Vision}, pages 14539--14548, 2021.

\bibitem{park2020video}
Minho Park, Sangmin Lee, and Yong~Man Ro.
\newblock Video frame interpolation via exceptional motion-aware synthesis.
\newblock In {\em ICASSP 2020-2020 IEEE International Conference on Acoustics,
  Speech and Signal Processing (ICASSP)}, pages 1958--1962. IEEE, 2020.

\bibitem{pathak2022fourcastnet}
Jaideep Pathak, Shashank Subramanian, Peter Harrington, Sanjeev Raja, Ashesh
  Chattopadhyay, Morteza Mardani, Thorsten Kurth, David Hall, Zongyi Li, Kamyar
  Azizzadenesheli, et~al.
\newblock Fourcastnet: A global data-driven high-resolution weather model using
  adaptive fourier neural operators.
\newblock {\em arXiv preprint arXiv:2202.11214}, 2022.

\bibitem{peleg2019net}
Tomer Peleg, Pablo Szekely, Doron Sabo, and Omry Sendik.
\newblock Im-net for high resolution video frame interpolation.
\newblock In {\em Proceedings of the IEEE/CVF Conference on Computer Vision and
  Pattern Recognition}, pages 2398--2407, 2019.

\bibitem{rahman2022u}
Md~Ashiqur Rahman, Zachary~E Ross, and Kamyar Azizzadenesheli.
\newblock U-no: U-shaped neural operators.
\newblock {\em arXiv preprint arXiv:2204.11127}, 2022.

\bibitem{raket2012motion}
Lars~Lau Rak{\^e}t, Lars Roholm, Andr{\'e}s Bruhn, and Joachim Weickert.
\newblock Motion compensated frame interpolation with a symmetric optical flow
  constraint.
\newblock In {\em International Symposium on Visual Computing}, pages 447--457.
  Springer, 2012.

\bibitem{shen2020blurry}
Wang Shen, Wenbo Bao, Guangtao Zhai, Li Chen, Xiongkuo Min, and Zhiyong Gao.
\newblock Blurry video frame interpolation.
\newblock In {\em Proceedings of the IEEE/CVF conference on computer vision and
  pattern recognition}, pages 5114--5123, 2020.

\bibitem{shi2021video}
Zhihao Shi, Xiaohong Liu, Kangdi Shi, Linhui Dai, and Jun Chen.
\newblock Video frame interpolation via generalized deformable convolution.
\newblock {\em IEEE Transactions on Multimedia}, 24:426--439, 2021.

\bibitem{shi2022video}
Zhihao Shi, Xiangyu Xu, Xiaohong Liu, Jun Chen, and Ming-Hsuan Yang.
\newblock Video frame interpolation transformer.
\newblock In {\em Proceedings of the IEEE/CVF Conference on Computer Vision and
  Pattern Recognition}, pages 17482--17491, 2022.

\bibitem{smolic2008intermediate}
Aljoscha Smolic, Karsten Muller, Kristina Dix, Philipp Merkle, Peter Kauff, and
  Thomas Wiegand.
\newblock Intermediate view interpolation based on multiview video plus depth
  for advanced 3d video systems.
\newblock In {\em 2008 15th IEEE International Conference on Image Processing},
  pages 2448--2451. IEEE, 2008.

\bibitem{tian2022video}
Haoyue Tian, Pan Gao, and Xiaojiang Peng.
\newblock Video frame interpolation based on deformable kernel region.
\newblock {\em arXiv preprint arXiv:2204.11396}, 2022.

\bibitem{tran2022video}
Quang~Nhat Tran and Shih-Hsuan Yang.
\newblock Video frame interpolation via down--up scale generative adversarial
  networks.
\newblock {\em Computer Vision and Image Understanding}, 220:103434, 2022.

\bibitem{wang2010depth}
Hung-Ming Wang, Chun-Hao Huang, and Jar-Ferr Yang.
\newblock Depth maps interpolation from existing pairs of keyframes and depth
  maps for 3d video generation.
\newblock In {\em Proceedings of 2010 IEEE International Symposium on Circuits
  and Systems}, pages 3248--3251. IEEE, 2010.

\bibitem{xue2019video}
Tianfan Xue, Baian Chen, Jiajun Wu, Donglai Wei, and William~T Freeman.
\newblock Video enhancement with task-oriented flow.
\newblock {\em International Journal of Computer Vision}, 127(8):1106--1125,
  2019.

\bibitem{zhang2018video}
Zhifeng Zhang, Li Song, Rang Xie, and Li Chen.
\newblock Video frame interpolation using recurrent convolutional layers.
\newblock In {\em 2018 IEEE Fourth International Conference on Multimedia Big
  Data (BigMM)}, pages 1--6. IEEE, 2018.

\end{thebibliography}
}

\end{document}